\pdfoutput=1
\documentclass[10pt,logo,copyright]{nvidiatechreport}
\usepackage[authoryear,sort&compress,round]{natbib}

\usepackage[utf8]{inputenc} % allow utf-8 input
\usepackage[T1]{fontenc}    % use 8-bit T1 fonts

\usepackage{parskip}        % no paragraph indents
\usepackage{url}            % simple URL typesetting
\usepackage{hyperref}       % hyperlinks
\usepackage{booktabs}       % professional-quality tables
\usepackage{amsfonts}       % blackboard math symbols
\usepackage{nicefrac}       % compact symbols for 1/2, etc.
\usepackage{microtype}      % microtypography
\usepackage{xcolor}         % colors
\usepackage[dvipsnames]{xcolor} % more color names
\usepackage{graphicx}
\usepackage{animate}        % for 360 video in teaser
\usepackage{subcaption}
\usepackage{tabularx}
\usepackage{makecell}
\usepackage{adjustbox}
\usepackage{setspace}
\newcolumntype{M}[1]{>{\centering\arraybackslash}m{#1}}
\usepackage{float}
\usepackage{tikz}
\usetikzlibrary{positioning,shapes,arrows}
\usepackage{amsmath,amsfonts,bm, bbm,leftindex}
\usepackage{multirow}
\usepackage{comment}
\usepackage{gensymb}
\usepackage{lipsum}
\usetikzlibrary{arrows.meta, positioning, fit}
\usepackage[para]{threeparttable}
\usepackage{tikz}
\usetikzlibrary{tikzmark}

\def\eqref#1{equation~\ref{#1}}
\def\1{\bm{1}}

\DeclareMathAlphabet{\mathsfit}{\encodingdefault}{\sfdefault}{m}{sl}
\SetMathAlphabet{\mathsfit}{bold}{\encodingdefault}{\sfdefault}{bx}{n}
\makeatletter
\let\save@mathaccent\mathaccent
\newcommand*\if@single[3]{%
  \setbox0\hbox{${\mathaccent"0362{#1}}^H$}%
  \setbox2\hbox{${\mathaccent"0362{\kern0pt#1}}^H$}%
  \ifdim\ht0=\ht2 #3\else #2\fi
  }
\newcommand*\rel@kern[1]{\kern#1\dimexpr\macc@kerna}
\newcommand*\widebar[1]{\@ifnextchar^{{\wide@bar{#1}{0}}}{\wide@bar{#1}{1}}}
\newcommand*\wide@bar[2]{\if@single{#1}{\wide@bar@{#1}{#2}{1}}{\wide@bar@{#1}{#2}{2}}}
\newcommand*\wide@bar@[3]{%
  \begingroup
  \def\mathaccent##1##2{%
    \let\mathaccent\save@mathaccent
    \if#32 \let\macc@nucleus\first@char \fi
    \setbox\z@\hbox{$\macc@style{\macc@nucleus}_{}$}%
    \setbox\tw@\hbox{$\macc@style{\macc@nucleus}{}_{}$}%
    \dimen@\wd\tw@
    \advance\dimen@-\wd\z@
    \divide\dimen@ 3
    \@tempdima\wd\tw@
    \advance\@tempdima-\scriptspace
    \divide\@tempdima 10
    \advance\dimen@-\@tempdima
    \ifdim\dimen@>\z@ \dimen@0pt\fi
    \rel@kern{0.6}\kern-\dimen@
    \if#31
      \overline{\rel@kern{-0.6}\kern\dimen@\macc@nucleus\rel@kern{0.4}\kern\dimen@}%
      \advance\dimen@0.4\dimexpr\macc@kerna
      \let\final@kern#2%
      \ifdim\dimen@<\z@ \let\final@kern1\fi
      \if\final@kern1 \kern-\dimen@\fi
    \else
      \overline{\rel@kern{-0.6}\kern\dimen@#1}%
    \fi
  }%
  \macc@depth\@ne
  \let\math@bgroup\@empty \let\math@egroup\macc@set@skewchar
  \mathsurround\z@ \frozen@everymath{\mathgroup\macc@group\relax}%
  \macc@set@skewchar\relax
  \let\mathaccentV\macc@nested@a
  \if#31
    \macc@nested@a\relax111{#1}%
  \else
    \def\gobble@till@marker##1\endmarker{}%
    \futurelet\first@char\gobble@till@marker#1\endmarker
    \ifcat\noexpand\first@char A\else
      \def\first@char{}%
    \fi
    \macc@nested@a\relax111{\first@char}%
  \fi
  \endgroup
}
\makeatother
\definecolor{darkred}{rgb}{0.7, 0.0, 0.0}

\usepackage{pifont}
\usepackage{wrapfig}
\usepackage{multicol}
\usepackage{multirow}
\usepackage{booktabs}
\usepackage{makecell}
\usepackage{xcolor}
\usepackage{tcolorbox}
\usepackage[table]{xcolor}

\usepackage[nameinlink]{cleveref}
\crefname{equation}{Eq.}{Eqs.}
\crefname{figure}{Fig.}{Figs.}
\crefname{section}{Sec.}{Sec.}
\crefname{appendix}{App.}{App.}
\crefname{table}{Tab.}{Tabs.}
\crefname{algorithm}{Algo}{Algo}
\crefname{thm}{Thm}{Thm}
\Crefname{thm}{Thm}{Thm}
\crefname{prop}{Prop}{Prop}
\usepackage{longtable}

\newcommand{\ours}{PhyCritic\xspace}

\definecolor{lightlightblue}{RGB}{235,245,255}

\newcommand{\crefnames}[3]{%
  \@for\next:=#1\do{%
    \expandafter\crefname\expandafter{\next}{#2}{#3}%
  }%
}

\title{PhyCritic: Multimodal Critic Models for Physical AI}

\author{Tianyi Xiong$^{1*}$, ~~Shihao Wang, ~~Guilin Liu, ~~Yi Dong, ~~Ming Li$^{1}$, ~~Heng Huang$^{1}$, ~~Jan Kautz, ~~Zhiding Yu$^\dag$ }

\begin{document}

\maketitle
\begin{abstract}

With the rapid development of large multimodal models, reliable judge and critic models have become essential for open-ended evaluation and preference alignment, providing pairwise preferences, numerical scores, and explanatory justifications for assessing model-generated responses. However, existing critics are primarily trained in general visual domains such as captioning or image question answering, leaving physical AI tasks involving perception, causal reasoning, and planning largely underexplored. We introduce \ours, a multimodal critic model optimized for physical AI through a two-stage RLVR pipeline: a physical skill warmup stage that enhances physically oriented perception and reasoning, followed by \emph{self-referential critic finetuning}, where the critic generates its own prediction as an internal reference before judging candidate responses, improving judgment stability and physical correctness. Across both physical and general-purpose multimodal judge benchmarks, \ours achieves strong performance gains over open-source baselines and, when applied as a policy model, further improves perception and reasoning in physically grounded tasks.

\end{abstract}\vspace{-2mm}  

\abscontent

\textbf{Links:} \hspace{2pt}
{
\hypersetup{urlcolor=nvidiagreen}
\href{https://research.nvidia.com/labs/lpr/phycritic/}{Project Page} 
% | \href{TODO}{HF Model} 
}

\begin{abstract}

With the rapid development of large multimodal models, reliable judge and critic models have become essential for open-ended evaluation and preference alignment, providing pairwise preferences, numerical scores, and explanatory justifications for assessing model-generated responses. However, existing critics are primarily trained in general visual domains such as captioning or image question answering, leaving physical AI tasks involving perception, causal reasoning, and planning largely underexplored. We introduce \ours, a multimodal critic model optimized for physical AI through a two-stage RLVR pipeline: a physical skill warmup stage that enhances physically oriented perception and reasoning, followed by \emph{self-referential critic finetuning}, where the critic generates its own prediction as an internal reference before judging candidate responses, improving judgment stability and physical correctness. Across both physical and general-purpose multimodal judge benchmarks, \ours achieves strong performance gains over open-source baselines and, when applied as a policy model, further improves perception and reasoning in physically grounded tasks.

\end{abstract}\vspace{-2mm}    
\section{Introduction}
\label{sec:intro}

Physical AI has emerged as a new frontier that involves visual perception, physical commonsense, spatial reasoning, and action-centric decision making within a single computational framework. Unlike traditional visual recognition tasks, physical AI requires a model to interpret complex multi-view observations, understand object affordances, reason over causal dynamics, and assess how hypothetical actions unfold in real environments. This paradigm spans 3D perception and spatial grounding \cite{zhang2025from}, robot-centric interaction understanding \cite{sermanet2024robovqa,walke2023bridgedata,wang2023holoassist,bu2025agibot}, and safety-critical domains such as autonomous driving \cite{marcu2024lingoqa,sima2024drivelm,nie2024reason2drive,xie2025vlms}. As these systems grow in scale and autonomy, the community increasingly relies on multimodal evaluation to measure whether a model’s reasoning is physically correct, visually grounded, and aligned with human expectations.

\begin{figure}[!ht]
    \centering
    \captionsetup{type=figure}
    % \vspace{-4mm}
    \includegraphics[trim=0 0 0 0, clip=True, width=0.99\textwidth]{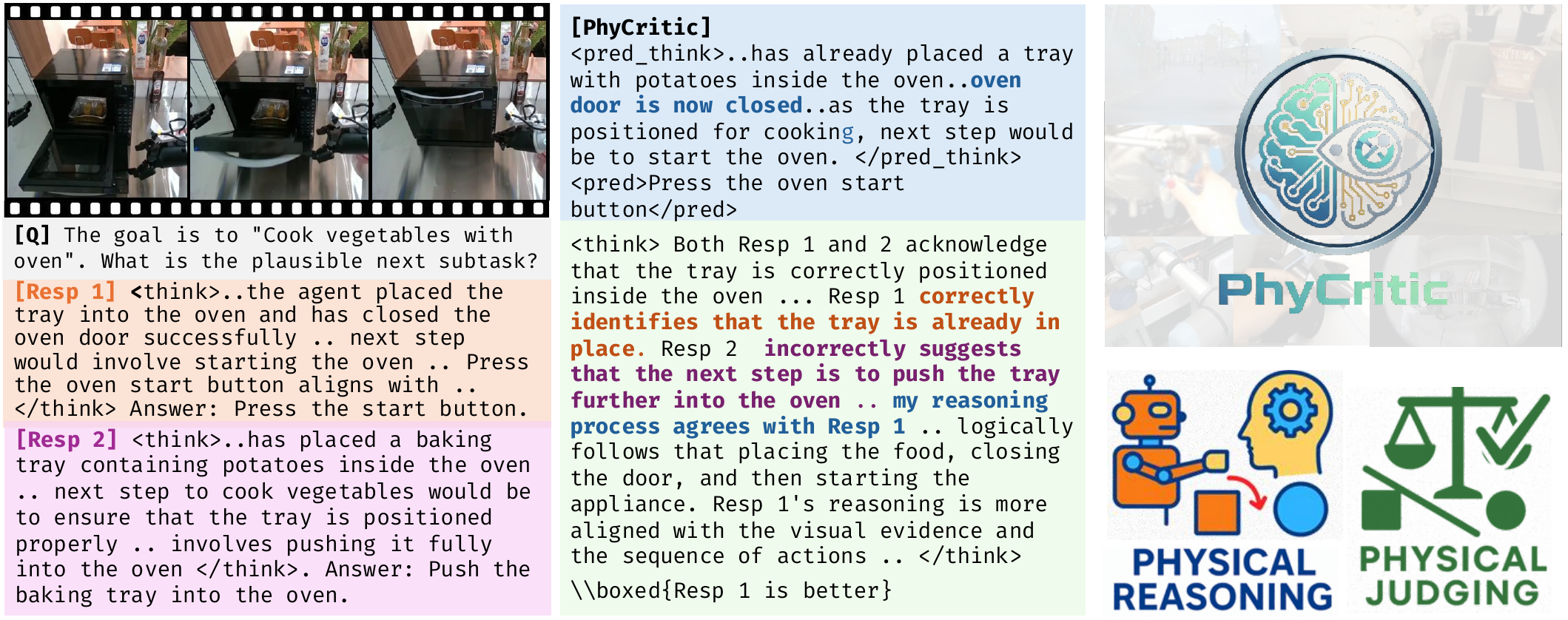}
    % \vspace{-1mm}
    \captionof{figure}{
    \ours first produces its own physics-aware reasoning and prediction, then explicitly applies it as reference in judging a pair of model responses. In this example, \ours first infers in its own prediction that ``the oven is closed". Based on this insight, the model then correctly identifies Response 1 as following the proper causal sequence while Response 2 proposes an unnecessary action. This self-referential process leads to more stable, physically correct judgments.
    }
    % \vspace{-1mm}
    \label{fig:teaser}
\end{figure}

However, despite rapid progress in multimodal large language models (MLLMs), the development of reliable multimodal \emph{critic models}---models that evaluate other models’ outputs---lags far behind. Existing reward or judge models focus predominantly on general domains such as captioning, STEM reasoning, and image question answering \cite{lee2024prometheus, xiong2025llava, wang2025llava, zhang2025r1,wang2025unified,unifiedreward-flex}. Yet, evaluation in physical AI is fundamentally different: the critic must assess whether reasoning is causally valid, whether a visual explanation adheres to actual physical configurations, and whether the final answer respects temporal, spatial, and dynamical constraints.

Recent works have started to extend multimodal judges and RL-based critic training \cite{wang2025llava, zhang2025r1, wang2025unified} to physical-related scenarios, while early efforts such as DriveCritic \cite{song2025drivecritic} underscore the importance of domain-specific judgment capabilities. But existing critics remain limited in three essential ways. (1) They lack \emph{physics awareness}, often failing to distinguish visually coherent but physically impossible reasoning. (2) Their training data focuses on broad multimodal evaluation rather than physically grounded scenarios involving manipulation, affordance reasoning, or embodied 3D interactions. (3) They do not ground their decisions in their own physical understanding of the question, which potentially leads to inconsistent or superficial verdicts.
In contrast, recent advances in reinforcement-finetuned multimodal \emph{policy} models show that RLVR-style verifiable rewards \cite{huang2025vision, liu2025noisyrollout, shen2025vlm, yu2025perception} and physically grounded reasoning datasets \cite{azzolini2025cosmos} can significantly improve multimodal reasoning and temporal consistency. Yet, these insights have not been systematically transferred to physical-related critic models, especially in settings where judgments must reflect physical truth rather than linguistic form.

\vspace{2mm}

\noindent\textbf{Our goal.}
We propose to bridge this gap by developing a new class of multimodal critics specifically designed for physical AI. Our model, called \textbf{\ours}, aims to evaluate multimodal responses involving physical perception, causal reasoning, and action or plan assessment, and to do so in a manner that is grounded, stable, and physically correct.

As shown in Fig.~\ref{fig:teaser}, \ours introduces the principle that a strong physical critic should behave like an expert human judge: before evaluating other models' responses, it should first solve the problem itself. This intuition motivates \emph{self-referential critic finetuning}, a two-stage reinforcement training framework:

\begin{itemize}[itemsep=3pt]
    \item \textit{Stage 1}: we apply standard GRPO on a small set of physical-related question–answer pairs to strengthen the model’s core physical perception and reasoning abilities, serving as a warm-up phase.
    \item \textit{Stage 2}: Building upon the Stage-1 warm-up, the critic is trained to (i) generate its own internal reasoning and prediction for the question, and then (ii) evaluate candidate responses with explicit reference to this self-prediction. Using GRPO with both critic and self-prediction rewards encourages stable critic behavior and coherent physics-aware reasoning.
\end{itemize}

\vspace{2mm}

\noindent\textbf{A new benchmark for physical critics.}
To rigorously evaluate judgment performance in physical contexts, we introduce \textbf{\ours-Bench}, a novel benchmark explicitly targeting multimodal critic models for physical AI. Built from diverse embodied datasets such as RoboVQA, BridgeData~V2, HoloAssist, and AgiBot World, \ours-Bench includes high-quality physical reasoning questions derived from Cosmos-Reason1 and paired candidate responses scored via verifiable ground truth. This enables fine-grained evaluation of reasoning correctness, visual grounding, and causal validity.

\vspace{2mm}

\noindent\textbf{Contributions.}
Our main contributions in this work are summarized as follows:
\begin{itemize}[itemsep=4pt]
\item We introduce a self-referential critic learning framework that explicitly grounds  the evaluation process in the model’s own physical perception and reasoning, implemented with a two-stage RLVR + GRPO pipeline.
\item We develop \textbf{\ours}, a multimodal critic specialized for assessing perception, causal reasoning, and planning in physical AI scenarios.
\item We construct a high-quality physical critic dataset spanning diverse embodied domains with paired candidate responses and verifiable preference labels.
\item We introduce \textbf{\ours-Bench}, a benchmark for evaluating multimodal critic models in physical contexts.
\item We demonstrate strong empirical gains across physical reasoning benchmarks (Cosmos-Reason1, CV-Bench, EgoPlan-Bench2) and general reward benchmarks (VL-RewardBench, Multimodal RewardBench), outperforming all open-source 7B/8B baselines.
\end{itemize}

Together, these results demonstrate that critic models benefit significantly from self-referential physical grounding, and that physical AI requires a new generation of physics-aware multimodal judge models.

\section{Related Works}

\noindent\textbf{VLMs for physical AI.}
Recent progress in vision–language models (VLMs) has expanded their scope from passive perception to physical AI tasks requiring perception, action, and spatial reasoning.
For spatial and 3D understanding, SPAR-7M and SPAR-Bench~\cite{zhang2025from} establish large-scale 3D-aware QA benchmarks to enhance multi-view and video-based grounding, while models such as 3D-LLM~\cite{hong20233d} and Point-LLM~\cite{guo2023point} integrate point-cloud features for open-vocabulary geometric reasoning.
In autonomous driving, benchmarks including Lingo-QA~\cite{marcu2024lingoqa}, DriveLM~\cite{sima2024drivelm}, Reason2Drive~\cite{nie2024reason2drive}, and DriveBench~\cite{xie2025vlms} assess multimodal reasoning, planning, and robustness under safety-critical conditions.
Vision–Language–Action (VLA) frameworks such as RT-2~\cite{zitkovich2023rt} and OpenVLA~\cite{kim2024openvla} unify perception, language, and control within a single policy space, while Open-X-Embodiment~\cite{o2024open} facilitates large-scale cross-robot generalization through shared representation learning.
Cosmos-Reason-1~\cite{azzolini2025cosmos} further introduces physically grounded chain-of-thought reasoning via supervised and reinforcement fine-tuning, bridging perception and planning for embodied agents.
Despite progress in perception and action, existing models often overlook the physical correctness and causal validity of their reasoning. \ours introduces a physics-aware critic for diagnostic evaluation in physical AI.

\begin{figure*}
    \centering
    \includegraphics[width=1.0\linewidth]{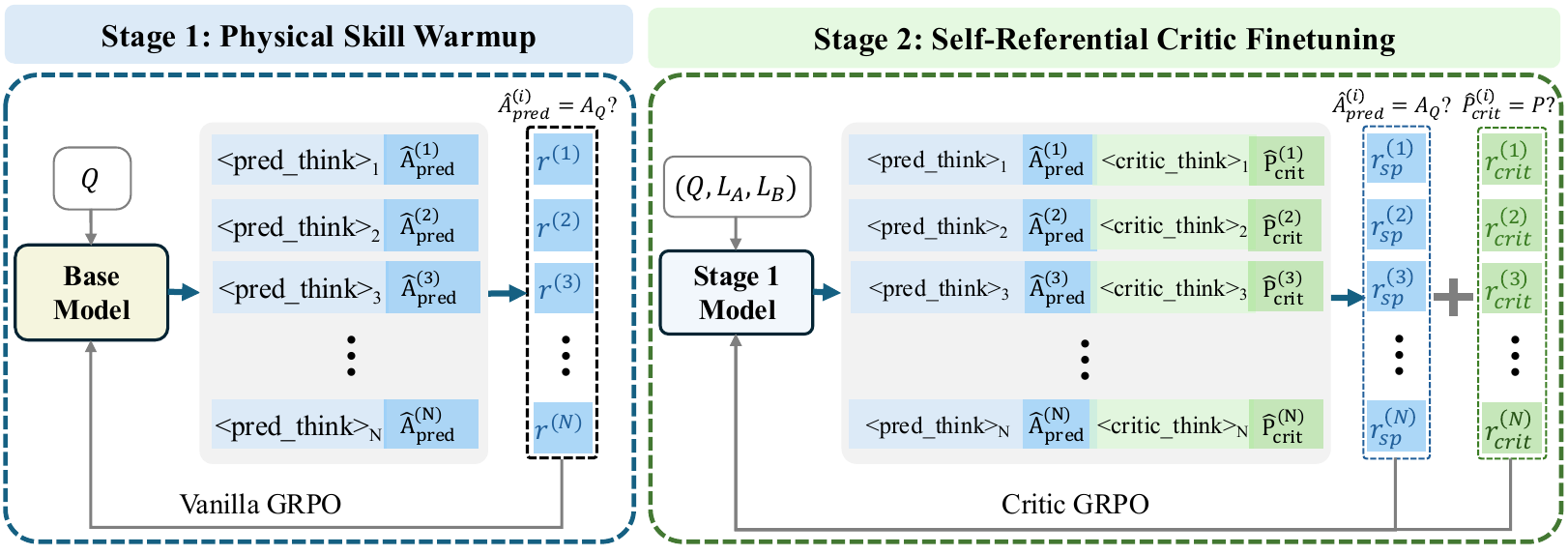}
    \vspace{-2mm}
    \caption{\textbf{\ours\ training pipeline.} We begin with GRPO training on physical-related QA pairs to enhance the VLM’s physical reasoning ability (\textit{left}), followed by self-referential critic finetuning to further develop its critique capacity (\textit{right}).}
    \label{fig:pipeline-phycrit}
\end{figure*}

% \vspace{2mm}

\noindent\textbf{Multimodal reward and critic models.} Vision-language models (VLMs) have recently been extended beyond perception and reasoning to act as evaluators, assessing the quality of model-generated responses. Early studies showed that proprietary models such as GPT-4V~\cite{zhang2023gpt} align closely with human judgments across diverse multimodal evaluation tasks.
Subsequent work fine-tuned open-source VLMs into reward or critic models, typically following two paradigms: (1) BT-style reward modeling with explicit scalar heads to predict preference scores~\cite{sun2023aligning, zang2025internlm, zhang2025basereward}, and (2) generative judges that produce textual reasoning and final verdicts in an autoregressive manner~\cite{lee2024prometheus, xiong2025llava, zhang2025r1, wang2025llava, wang2025unifiedthink}, or combine both approaches~\cite{zhang2025mm}.
Later extensions incorporated critic training with reinforcement learning~\cite{zhang2025r1, wang2025llava, wang2025unifiedthink} to enhance judgment accuracy.
However, most efforts remain focused on general visual domains such as captioning, STEM reasoning, or visual question answering, leaving physical AI scenarios underexplored.
Among earlier attempts, WorldModelBench~\cite{li2025worldmodelbench} evaluates video-generation models on instruction following and physics adherence, underscoring the need for physics-aware multimodal evaluation.
More recently, DriveCritic~\cite{song2025drivecritic} employed LMM-based critics to assess trajectory pairs in autonomous driving, yet its scope remains and does not apply to diverse, grounded multimodal text responses. 
Our work broadens the critic capabilities of open-source VLMs toward general physical AI domains, emphasizing evaluation settings that involve physical perception, reasoning, and planning.

% \vspace{2mm}

\noindent\textbf{Multimodal reinforcement finetuning.} 
Reinforcement finetuning has proven effective in enhancing the capabilities of vision-language models (VLMs). Early explorations applied reinforcement learning with human feedback (RLHF~\cite{ouyang2022training}) and AI feedback (RLAIF~\cite{lee2023rlaif}) paradigms, typically using PPO~\cite{schulman2017proximal} or DPO~\cite{rafailov2023direct}, to mitigate hallucinations~\cite{sun2023aligning, yu2024rlhf, yu2025rlaif} and improve video understanding~\cite{zhang2025direct}. Following the success of DeepSeek-R1~\cite{guo2025deepseek}, subsequent studies have increasingly adopted Reinforcement Learning with Verifiable Rewards (RLVR) approaches to strengthen visual reasoning in multimodal mathematics and science~\cite{huang2025vision, liu2025modality, yang2025r1, meng2025mm, wang2025sota, liu2025noisyrollout}, perception-related tasks~\cite{yu2025perception, zhou2025r1, shen2025vlm}, video reasoning~\cite{feng2025video}, and domain-specific applications including medical imaging~\cite{lai2025med}, GUI agents~\cite{luo2025gui}, and robotic reasoning~\cite{azzolini2025cosmos}.
These efforts primarily target tasks with verifiable ground-truth supervision, aiming to improve models as policy for problem solving. 
More recently, RLVR-based methods have been extended to multimodal critic finetuning—training models to evaluate rather than generate responses—thereby enhancing judgment accuracy on image-grounded tasks~\cite{zhang2025r1, wang2025llava, wang2025unified,lithink}.
\ours extends multimodal RLVR to develop physical critics that evaluate responses involving perception, action, and planning in physical contexts, while enhancing the model’s ability for physical reasoning.

\section{\ours: Physical Critic for VLMs}

Inspired by human critical thinking~\cite{lai2011critical}, we identify two key principles for consolidating physics-related critic capacities in VLMs. (1) These capacities should be developed from model’s intrinsic ability for physics-related perception and reasoning. (2) Building on the success of multimodal critics in both general image understanding~\cite{xiong2025llava, lee2024prometheus, wang2025llava, zhang2025r1} and visual generation domains~\cite{liu2025improving, wang2025unified}, critic performance should be consolidated through explicit critic training with strong preference signals.
Therfore, \ours adopts a two-stage reinforcement fine-tuning pipeline. Starting from a base model, we first fine-tune it on a small set of physical-related question-answer pairs to enhance its physical perception and reasoning. We then introduce \emph{self-referential critic fine-tuning}, a reinforcement learning stage where the critic is prompted to generate its own prediction for the user’s query before evaluating pairs of model responses. An overview is provided in Fig.~\ref{fig:pipeline-phycrit}.

\subsection{Task Formulation}
To implement this mechanism, our training data are organized as tuples $(Q, L_A, L_B, A_Q, P)$, where:
\begin{itemize}
\item $Q$ denotes a multimodal prompt consisting of a user question together with visual inputs;
\item $L_A$ and $L_B$ are two candidate responses to be evaluated;
\item $A_Q$ denotes the ground-truth answer to $Q$;
\item $P \in {A, B}$ is a binary preference label indicating which response demonstrates higher quality.
\end{itemize}
\subsection{Two-Stage Training Pipeline}

As existing VLMs are primarily trained on general image–text pairs with limited exposure to physically grounded data, we devise a two-stage training pipeline to enhance physical critic capacity.

% % \vspace{2mm}

\noindent\textbf{Stage 1: Physical Skill Warmup with RLVR.}
Before introducing the complex critic task, we first ensure that the model possesses solid foundational perception and reasoning capabilities within the physical domain. In this stage, we use a verifiable dataset containing only $(Q, A_{Q})$ pairs and adopt a standard reinforcement learning setup with accuracy reward
\[
    r = \mathbb{I} (\hat{A}_{pred}(Q) = A_{Q})
\]
The objective is to align the model to produce accurate and reliable predictions $\hat{A}_{\mathrm{pred}}$, establishing a strong foundation for critic training in the next stage.

% \vspace{2mm}
% 
\noindent\textbf{Stage 2: Self-Referential Critic Finetuning.}
After acquiring foundational physical capacities, we utilize the complete dataset $(Q, L_A, L_B, A_{\mathrm{Q}}, P)$ to align the model’s judgment over a pair of responses, $\hat{P}_{\mathrm{pred}}(Q, L_A, L_B)$, with the ground-truth preference label $P$. As will be illustrated in Section~\ref{self-ref}, this stage encourages the model to anchor its external critic judgment to its internal self-prediction $\hat{A}_{\mathrm{pred}}$, thereby promoting the development of more grounded and interpretable physical critic capacities.

\subsection{Self-Referential Critic Finetuning}
\label{self-ref}

\definecolor{lightlightgreen}{RGB}{210, 245, 210}  % very soft green

\begin{table*}[!ht]
\centering
\begin{minipage}{1.0\textwidth}
\vspace{0mm}    
\centering
\begin{tcolorbox}
\small
\renewcommand{\arraystretch}{0.96}  %
\begin{tabular}{p{0.96\textwidth}}
You are an expert in evaluating responses from multimodal models. Your task is to compare two model responses to the same question grounded on the provided image or video. You should assess both the \textbf{quality of the reasoning process} and the \textbf{factual correctness of the final answer} in each response. \\[4pt]

Your evaluation should focus on two main aspects: \\[2pt]

1. \textbf{Reasoning Process Quality:} \\
Evaluate the model’s reasoning based on the following aspects: \\
-- \textit{Truthfulness:} All factual statements must be correct and verifiable. No fabricated details or contradictions with the visual input; 
-- \textit{Visual Groundedness:} The reasoning must accurately reference and interpret visual elements from the provided image or video;
-- \textit{Logical Validity:} The reasoning should follow a logically consistent, step-by-step progression; 
-- \textit{Efficiency and Clarity:} Reasoning should be clear, purposeful, and focused. Thoughtful self-correction is acceptable, but avoid unnecessary repetition or irrelevant information. \\[8pt]

2. \textbf{Final Answer Accuracy:} \\
Evaluate whether the final answer is factually correct given the question and the visual input. \\[2pt]

Here are the inputs for evaluation: \\
\texttt{[Question]:\{question\}}; \
\texttt{[Response 1]:\{resp1\}}; \
\texttt{[Response 2]:\{resp2\}} \\[4pt]

To begin your evaluation, first generate your own reasoning process towards solving the question, enclosing it within \texttt{<pred\_think></pred\_think>}. Then, provide your own answer to the question in \texttt{<pred></pred>}. After that, provide a detailed justification comparing the two responses, using your own response as a reference point. Evaluate each response according to the criteria above, considering both reasoning quality and final answer correctness. Your explanation must clearly reference specific details from both responses, your own response, and the visual evidence. % \\[4pt]
Finally, make a clear and strict decision on which response is better overall. Output your decision in one of the following formats: \\
-- Response 1 is better; 
-- Response 2 is better \\[4pt]

\rowcolor{lightlightgreen}
You FIRST write your reasoning process to the question in \texttt{<pred\_think></pred\_think>}, THEN generate your own answer in \texttt{<pred></pred>}, THEN think about the comparison and judgment process as an internal monologue (enclosed within \texttt{<think></think>}), and FINALLY provide the final answer in \texttt{\textbackslash boxed\{\}}. \\
\end{tabular}
\renewcommand{\arraystretch}{1.0}  % 
\end{tcolorbox}
\vspace{-2mm}
\caption{\textbf{Critic prompt for self-referential critic fine-tuning.} 
After presenting detailed evaluation criteria, it explicitly instructs the judge model to first generate its own reasoning and prediction for the given question, then use its self-prediction as a reference during the critique process. 
\colorbox{lightlightgreen}{Format prompt} regions are highlighted in green.}
\vspace{-3mm}
\label{tab:critic_prompt_self_reference}
\end{minipage}
\end{table*}

Our core idea is that a critic's judgment should be grounded in its own understanding of the problem. Therefore, during training, the model $\pi_{\theta}$ is required to perform two tasks concurrently:

\begin{enumerate}[itemsep=2pt, topsep=4pt]
    \item \textbf{Self-Prediction:} The model first generates its own internal prediction $\hat{A}_{pred}$ in response to the prompt $Q$.
    
    \item \textbf{Preference Judgement:} 
    Subsequently, acting as a critic, the model produces its preference prediction $\hat{P}_{\mathrm{pred}}$ for the context $(Q, L_A, L_B)$, while being explicitly instructed to ground its evaluation process on the previously generated self-prediction $\hat{A}_{\mathrm{pred}}$.

\end{enumerate}

\noindent\textbf{Reward Design.}
The total reward $r_{\mathrm{total}}$ is composed of an accuracy reward $r_{acc}$ and a format reward $r_{format}$.
\[
    r_{total}= r_{acc}+ r_{form} * \alpha_{form}
\]
where $\alpha_{\mathrm{form}}$ is a weighting coefficient for format reward.

% \vspace{2mm}

\noindent\textbf{Accuracy Reward.}
In self-referential critic fine-tuning, the accuracy reward $r_{\mathrm{acc}}$ consists of two components:

\begin{itemize}[itemsep=2pt, topsep=4pt]
    \item \textbf{Self-Prediction Reward ($r_{\mathrm{sp}}$):}
    Evaluates the correctness of the model’s internal knowledge by comparing its self-prediction $\hat{A}_{\mathrm{pred}}$ with the ground-truth answer $A_{\mathrm{Q}}$:
    \[
    r_{\mathrm{sp}} = \mathbb{I}(\hat{A}_{\mathrm{pred}} = A_{\mathrm{Q}})
    \]
    which encourages the VLM to first become a reliable problem solver.

    \item \textbf{Critic Reward ($r_{\mathrm{crit}}$):}
    Evaluates the model’s judgment capability as a critic by checking whether its predicted preference $\hat{P}_{\mathrm{crit}}$ matches the golden preference $P$:
    \[
    r_{\mathrm{crit}}  = \mathbb{I}(\hat{P}_{\mathrm{crit}}(Q, L_A, L_B) = P)
    \]

\end{itemize}

The overall accuracy reward $r_{\mathrm{acc}}$ is computed as a weighted sum of the two components:
\[
r_{\mathrm{acc}} = \alpha_{\mathrm{sp}} r_{\mathrm{sp}} + \alpha_{\mathrm{crit}} r_{\mathrm{crit}}
\]
where $\alpha_{\mathrm{sp}}$ and $\alpha_{\mathrm{crit}}$ are weighting coefficients controlling their relative importance.

% \vspace{2mm}
\noindent\textbf{Format Reward ($r_{\mathrm{form}}$).}
This reward evaluates whether the critic’s output follows the predefined self-referential structure:
\[
r_{\mathrm{form}} =
\begin{cases}
1.0 & \text{if } \texttt{<pred\_think>}, \texttt{<pred>}, \\
 &  \texttt{<think>} \text{ and } \texttt{\textbackslash boxed\{\}} \text{ exist},\\[3pt]
0.5 & \text{if only } \texttt{<think>} \text{ and } \texttt{\textbackslash boxed\{\}} \text{ exist},\\[3pt]
0 & \text{otherwise.}
\end{cases}
\]

\vspace{2mm}

\noindent\textbf{Optimization Algorithm.} Inspired by the strong performance DeepSeek-R1~\cite{guo2025deepseek},  we employ Group Relative Policy Optimization (GRPO) as the optimization algorithm. GRPO is a policy-gradient based reinforcement-learning method originally introduced in the context of  mathematical reasoning~\cite{shao2024deepseekmath}. It omits the need for a learned value-network in PPO, and instead computes advantages by comparing multiple sampled trajectories as a group.
\begin{equation*}
\begin{aligned}
\mathcal{L}_{\text{GRPO}}\ &
\mathbb{E}_{o\sim\pi_\theta}\big[
\min(\rho_{o}A_{o}, \text{clip}(\rho_o,1-\epsilon,1+\epsilon)A_o)
\big] \\
&\quad - \beta\, \mathcal{D}_{\mathrm{KL}}(\pi_\theta \Vert \pi_{\mathrm{ref}})
\end{aligned}
\end{equation*}
where $\rho_o = \frac{\pi_\theta(o)}{\pi_{ref}(o)}$ is the probability ratio, $A_o = \frac{r_o - \bar{r}}{\text{std}(r)}$ is the group-relative advantage, computed by normalizing each outcome reward $r_o$ for the sampled trace $o$.

\vspace{2mm}

\noindent\textbf{Critic Prompt.}
As shown in Tab.~\ref{tab:critic_prompt_self_reference}, we design a rigorous critic prompt template that explicitly instructs the model to follow a clear set of predefined evaluation criteria, and to reference its critique process with respect to its own prediction. Following DeepSeek-R1~\cite{guo2025deepseek}, we append a format prompt at the end to standardize the model’s reasoning behavior and output structure.

\subsection{Critic Training Dataset}

To enhance critic capacities, we construct a dataset spanning diverse physical scenarios and video domains, paired with high-quality questions probing perception, reasoning, and causality. Candidate responses of varying quality enable the critic to learn nuanced distinctions in reasoning accuracy and visual grounding.

We select videos from four major robotics and embodied datasets—RoboVQA~\cite{sermanet2024robovqa}, BridgeData V2~\cite{walke2023bridgedata}, HoloAssist~\cite{wang2023holoassist}, and AgiBot World~\cite{bu2025agibot}—covering both egocentric and third-person views, a wide range of manipulation behaviors (e.g., grasping, stacking, folding, assembling), and diverse physical contexts from simulated kitchens to real-world workspaces. For questions, we build upon the Cosmos-Reason1 RL dataset~\cite{azzolini2025cosmos}, which provides 800 high-quality question–answer pairs requiring rich physical perception, planning, and reasoning capacities.

Model outputs are collected from seven multimodal models spanning proprietary (GPT-4o~\cite{hurst2024gpt}, Gemini-2.5-Flash~\cite{comanici2025gemini}), open-source (Qwen2.5-VL-72B~\cite{bai2025qwen2}, InternVL3-38B~\cite{zhu2025internvl3}), and RL-finetuned systems with enhanced long-horizon reasoning (Cosmos-Reason1-7B~\cite{azzolini2025cosmos}, Video-R1~\cite{feng2025video}, MiMo-VL-7B~\cite{team2025kimi}). All models are prompted in a Chain-of-Thought (CoT) manner to produce explicit reasoning traces with their final responses.

For ground-truth critic preference, we apply a simple accuracy-based labeling: GPT-4o verifies each response against the ground-truth answer and assigns a binary score (1 = preferred, 0 = rejected). We then construct response pairs, each consisting of one chosen and one rejected response. After balancing response lengths and distribution, we obtain a final critic dataset of 3,258 samples.

\section{\ours-Bench}

\begin{wrapfigure}{R}{0.57\textwidth}
    \centering
    % \vspace{-3mm}
    \includegraphics[width=1\linewidth]{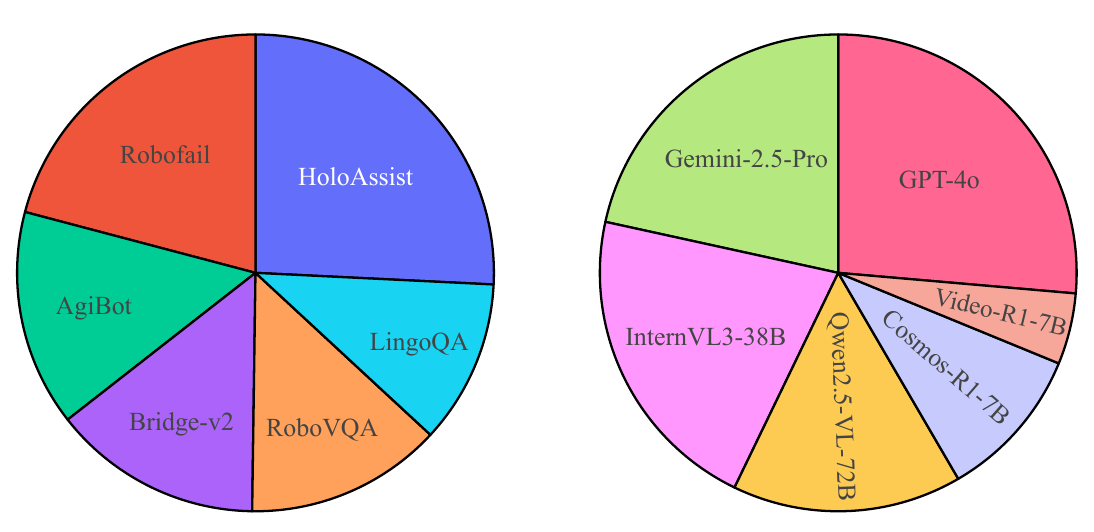}
    \caption{
    Distribution of prompt sources (\textit{left}) and model responses (\textit{right}) in \ours-Bench.
    }
    \vspace{-5mm}
    \label{fig:bench_statistics}
\end{wrapfigure}
Existing multimodal reward benchmarks primiarily cover prompts from general visual domains such as captioning~\cite{ye2025painting}, STEM reasoning~\cite{zhang2025mm, chen2024mllm,xiong2025multicritbenchmarkingmultimodaljudges} and general image question answering~\cite{li2025vl, yasunaga2025multimodal}, does not cover physical AI domains that involves judgments related to physical perception, actions and planning. We thus introduce \ours-Bench, a fine-grained and challeging benchmark targeted in physical AI domains for multimodal judge models.

\begin{table*}[t]
\centering
\small
\setlength{\tabcolsep}{2.3pt}
\scalebox{0.975}{
\begin{tabular}{l|ccccccc|ccccc|c}
\toprule
\multirow{2}{*}{Model} 
& \multicolumn{7}{c|}{\ours-Bench} 
& \multicolumn{5}{c|}{VL-RewardBench} % {VL-RewardBench~\cite{li2025vl}} 
& \multicolumn{1}{c}{MM-RB.} % {MM-RB.~\cite{yasunaga2025multimodal}} 
\\
\cmidrule(lr){2-8} \cmidrule(lr){9-13} \cmidrule(lr){14-14}
 & overall & AgiB. & Holo. & R-VQA & Bridge. & R-Fail & Lingo.
 & overall & macro & General & Hallu. & Reason 
 & overall \\
\midrule

\textcolor{gray}{Gemini-2.5-Pro}
%~\cite{comanici2025gemini}   
& \textcolor{gray}{78.2} & \textcolor{gray}{87.9} & \textcolor{gray}{82.8} & \textcolor{gray}{86.7} & \textcolor{gray}{68.8} & \textcolor{gray}{76.6} & \textcolor{gray}{60.0}
& \textcolor{gray}{74.9} & \textcolor{gray}{72.9} & \textcolor{gray}{59.1} & \textcolor{gray}{85.2} & \textcolor{gray}{74.4} & \textcolor{gray}{85.4} \\

\textcolor{gray}{Gemini-2.5-Flash}
%~\cite{comanici2025gemini} 
&  \textcolor{gray}{67.1} & \textcolor{gray}{63.6} & \textcolor{gray}{70.7} & \textcolor{gray}{73.3} & \textcolor{gray}{81.2} & \textcolor{gray}{63.8} & \textcolor{gray}{44.0}
& \textcolor{gray}{71.9} & \textcolor{gray}{72.0} & \textcolor{gray}{58.0} & \textcolor{gray}{77.0} & \textcolor{gray}{81.0} & \textcolor{gray}{82.5} \\

\textcolor{gray}{GPT-4o}
% ~\cite{hurst2024gpt}           
& \textcolor{gray}{64.7} & \textcolor{gray}{57.6} & \textcolor{gray}{70.2} & \textcolor{gray}{46.7} & \textcolor{gray}{75.0} & \textcolor{gray}{72.3} & \textcolor{gray}{56.0} & 
\textcolor{gray}{65.8} & \textcolor{gray}{62.4} & \textcolor{gray}{49.6} & \textcolor{gray}{67.6} & \textcolor{gray}{70.5} & \textcolor{gray}{71.5} \\
\midrule
\multicolumn{13}{l}{\textit{Open-Source 7B/8B VLMs and Physical-Task–Tuned Variants}}\\
\midrule
Eagle-2.5-8B
%~\cite{chen2025eagle} 
& \underline{56.0} & \underline{75.8} & 50.0 & \underline{60.0} & 50.0 & 48.9 & \textbf{60.0} &
 50.2 & 49.7 & \underline{41.4} & 48.6 & 59.3 & \underline{64.4} \\

Qwen2.5-VL-7B
% ~\cite{bai2025qwen2} 
& 51.6 & 51.5 & 51.7 & 46.7 & 50.0 & \textbf{57.4} & 48.0
                 & \underline{53.2} & \underline{50.9} & 40.9 & 54.3 & 57.4 & 64.0 \\

Robobrain2.0-7B
% ~\cite{team2025robobrain} 
& 54.7 & 51.5 & \textbf{65.5} & 43.3 & 46.9 & \underline{55.3} & \underline{56.0} &42.4  &44.2 & 39.2& \underline{55.8} & 37.5&  50.5 \\
% \hline
Cosmos-R1-7B
% ~\cite{azzolini2025cosmos} 
& 51.1 & 36.4 & \underline{53.4} & 43.3 & \textbf{68.8} & 51.1 & 52.0 &  44.8  & 44.8 & 33.1 & 41.3 & \underline{59.9} & 54.8 \\
\rowcolor{lightlightblue}
\ours-7B (ours) & \textbf{68.0} & \textbf{78.8} & \textbf{65.5} & \textbf{86.7} & \underline{65.6} & \textbf{57.4} & \textbf{60.0}
                 & \textbf{57.3} & \textbf{54.9} & \textbf{45.3} & \textbf{58.6} & \textbf{60.9} & \textbf{65.9} \\

\bottomrule
\end{tabular}
}
\caption{Performance comparison on \ours-Bench and two general multimodal reward benchmarks. Best and second-best results of \textit{open-source} VLMs are marked in \textbf{bold} and \underline{underline}, respectively. *MM-RB. is an abbreviated form of Multimodal RewardBench~\cite{yasunaga2025multimodal}.  }
\label{tab:judgebench-bench}
\end{table*}

\ours-Bench primarily comprises 225 evaluation samples covers evaluations in two physical-AI scenarios:
1) \textit{Robotics tasks} – videos from RoboVQA~\cite{sermanet2024robovqa}, BridgeData V2~\cite{walke2023bridgedata}, HoloAssist~\cite{wang2023holoassist}, AgiBot~\cite{bu2025agibot}, and RoboFail~\cite{liu2023reflect}, with questions adapted from CosmosReason1-Bench~\cite{azzolini2025cosmos};
2) \textit{Autonomous driving} – videos and action- or prediction-related questions from LingoQA~\cite{marcu2024lingoqa}.
Model responses are collected from seven VLMs spanning proprietary, open-source, and RL-finetuned families. An overview of the prompt and response model statistics is shown in Fig.~\ref{fig:bench_statistics}.

Following prior works, we formulate each evaluation instance in \ours-Bench as a pairwise preference tuple $(x, l_a, l_b, p)$, where $q$ is the multimodal prompt containing the visual input and user query, $l_a$ and $l_b$ are candidate responses, and $p \in {a,b}$ is the ground-truth preference label. For response-pair and preference-label construction, we follow the JudgeBench~\cite{tan2024judgebench} pipeline. Each prompt is assigned to one model from a pool, which generates $N{=}8$ CoT responses via temperature sampling. GPT-4o verifies each response against the ground-truth answer, identifying correct and incorrect predictions. The final pair comprises one correct and one incorrect response, with correctness serving as the preference label. During evaluation, each VLM acts as a pairwise judge, and performance is measured by its preference consistency with the ground truth:
\[
    Acc = \mathbb{I}\big{(}\text{VLM}(q,l_a,l_b) = p\big{)}
\]

\section{Experiments}

\noindent\textbf{Implementation Details.} We adopt Qwen2.5-VL-7B-Instruct~\cite{bai2025qwen2} as our base model and the veRL framework for RL finetuning. In the warmup stage, we train the model using vanilla GRPO on the Cosmos-Reason1-RL dataset~\cite{azzolini2025cosmos} for 80 steps. In the critic finetuning stage, we perform self-referential critic finetuning for 300 steps. For both stages, we use a batch size of 128, a learning rate of $1\times10^{-6}$, and set the KL coefficient to 0.01. For critic finetuning, we set $\alpha_{sp} = 0.2, \alpha_{crit} = 0.7$ and $\alpha_{form} = 0.1$.

% % \vspace{2mm}

\noindent\textbf{Evaluation Benchmark.} 
We evaluate our approach on both critic performance and its capacity for handling Physical-AI tasks.
For critic performance, we use our curated \ours-Bench for physical-related critique, together with two general reward benchmarks—VL-RewardBench~\cite{li2025vl} and Multimodal-RewardBench~\cite{yasunaga2025multimodal}—to evaluate critic capacity in general visual understanding and knowledge-intensive reasoning.
For physical capacity, we evaluate our methods on three physical-related multimodal benchmarks. CosmosReason1-Bench probes physical commonsense and causal/affordance reasoning across agility, holistic perception, robot-VQA, tool-use/bridging, and failure-robustness sub-tasks~\cite{azzolini2025cosmos}. CV-Bench assesses whether visual representations support downstream control through 2D/3D spatial cognition and cross-view consistency~\cite{tong2024cambrian}. EgoPlanBench2 targets egocentric daily-task planning (Daily/Hobbies/Recreation/Work), measuring the full perception–decomposition–planning pipeline~\cite{qiu2024egoplan}.

% \vspace{2mm}

\noindent\textbf{Baselines.}
We compare \ours against two groups of baselines.
(1) \textit{General-purpose VLMs}: leading open-source models trained for visual instruction following, including Qwen2.5-VL-7B~\cite{bai2025qwen2} and Eagle2.5-8B~\cite{chen2025eagle}.
(2) \textit{Physical reasoning–oriented VLMs} — two representative lines built upon the same Qwen2.5-VL backbone as ours: Cosmos-Reason1~\cite{azzolini2025cosmos}, optimized for physical reasoning through large-scale supervised finetuning on millions of physical common-sense and embodied reasoning samples distilled from DeepSeek-R1 traces, followed by reinforcement finetuning on object interaction and temporal-consistency tasks; and RoboBrain2~\cite{team2025robobrain}, a perception-to-action model trained via multi-stage supervised and reinforcement finetuning that emphasizes task decomposition, subgoal generation, and robustness in embodied environments.

\definecolor{lightlightgray}{gray}{0.92}

% \definecolor{lightlightblue}{RGB}{235,245,255}
\begin{table*}[t]
\centering
\small
\setlength{\tabcolsep}{1.4pt}
\scalebox{0.97}{
\begin{tabular}{l|cccccc|ccc|ccccc}
\toprule
\multirow{2}{*}{Model} 
& \multicolumn{6}{c|}{CosmosReason1-Bench} % {CosmosReason1-Bench~\cite{azzolini2025cosmos}} 
& \multicolumn{3}{c|}{CV-Bench} % {CV-Bench~\cite{tong2024cambrian}} 
& \multicolumn{5}{c}{EgoPlanBench2} % {EgoPlanBench2~\cite{yasunaga2025multimodal}} 
\\
\cmidrule(lr){2-7} \cmidrule(lr){8-10} \cmidrule(lr){11-15}
 & overall & AgiB. & Holo. & R-VQA & Bridge. & RoboFail 
 & overall & 2D & 3D & overall & Daily & Hobb. & Recreat. & Work\\
\midrule

\textcolor{gray}{Gemini-2.5-Pro}
%~\cite{comanici2025gemini}  
& \textcolor{gray}{64.7} & \textcolor{gray}{45.0} & \textcolor{gray}{84.0} & \textcolor{gray}{83.6} & \textcolor{gray}{45.0} & \textcolor{gray}{74.0}  & \textcolor{gray}{85.4} & \textcolor{gray}{79.3} & \textcolor{gray}{91.5} & \textcolor{gray}{42.8} & \textcolor{gray}{44.2} & \textcolor{gray}{43.1} & \textcolor{gray}{46.4} & \textcolor{gray}{39.6} \\
\textcolor{gray}{Gemini-2.5-Flash}
%~\cite{comanici2025gemini} 
& \textcolor{gray}{57.9} & \textcolor{gray}{42.7} & \textcolor{gray}{68.2} & \textcolor{gray}{78.2} & \textcolor{gray}{35.5} & \textcolor{gray}{62.8} & \textcolor{gray}{74.0} & \textcolor{gray}{74.4} & \textcolor{gray}{73.6} & \textcolor{gray}{37.1} & \textcolor{gray}{38.7} & \textcolor{gray}{35.6} & \textcolor{gray}{43.7} & \textcolor{gray}{33.4}  \\
\textcolor{gray}{GPT-4o}
%~\cite{hurst2024gpt} 
& \textcolor{gray}{56.3} & \textcolor{gray}{38.0} & \textcolor{gray}{64.0} & \textcolor{gray}{78.2} & \textcolor{gray}{33.0} & \textcolor{gray}{66.0}  & \textcolor{gray}{79.1} & \textcolor{gray}{73.2} & \textcolor{gray}{85.0} & \textcolor{gray}{41.8} & \textcolor{gray}{47.4} & \textcolor{gray}{40.0} & \textcolor{gray}{44.8} & \textcolor{gray}{35.6} \\

% \hline
\midrule
\multicolumn{13}{l}{\textit{Open-Source 7B/8B VLMs and Physical-Task–Tuned Variants}}\\
% \hline
\midrule

Eagle-2.5-8B
%~\cite{chen2025eagle}
& 54.3 & 39.0 & \underline{67.0} & 67.4 & 29.0 & \textbf{68.0} &  77.9   & \textbf{77.5}    & 78.3   & \textbf{43.0}    & \textbf{49.4}    & \underline{41.7} & \underline{47.0} & \underline{35.1} \\
Qwen2.5-VL-7B
% ~\cite{bai2025qwen2}
& 54.3 & \hphantom{$^{\dagger}$}40.4$^{\dagger}$ & \hphantom{$^{\dagger}$}50.0$^{\dagger}$ & \hphantom{$^{\dagger}$}82.5$^{\dagger}$ & \hphantom{$^{\dagger}$}\underline{38.0$^{\dagger}$} & \hphantom{$^{\dagger}$}57.6$^{\dagger}$ & 78.9 & 75.1 & \underline{82.7} & 32.6 & 36.7 & 27.7 & 33.9 & 32.5\\
Robobrain-2.0-7B
% ~\cite{team2025robobrain} 
&  50.4 &37.0 &41.0 & \underline{85.4}& 29.0& 56.0 & \textbf{\hphantom{$^{\dagger}$}85.8$^{\dagger}$} & -- & -- & \hphantom{$^{\dagger}$}33.2$^{\dagger}$ & \hphantom{$^{\dagger}$}39.4$^{\dagger}$ & \hphantom{$^{\dagger}$}32.2$^{\dagger}$ & \hphantom{$^{\dagger}$}33.9$^{\dagger}$ & \hphantom{$^{\dagger}$}29.7$^{\dagger}$ \\
Cosmos-R1-7B
& \underline{63.0}    & \hphantom{$^{\dagger}$}\underline{49.4}$^{\dagger}$   & \hphantom{$^{\dagger}$}63.0$^{\dagger}$  & \hphantom{$^{\dagger}$}83.8$^{\dagger}$   & \hphantom{$^{\dagger}$}\textbf{58.8}$^{\dagger}$  & \hphantom{$^{\dagger}$}60.0$^{\dagger}$
    & 75.2    & 74.9    & 75.7    & 29.8  & 33.5    &  31.2 & 30.6 & 24.5\\
\rowcolor{lightlightblue}
\ours-7B (ours)       & \textbf{63.9} & \textbf{53.0} & \textbf{68.0} & \textbf{95.5} & 37.0 & \underline{63.0}
                  & \underline{79.7}    & \underline{75.5}    & \textbf{83.9}    & \underline{42.3} & \underline{45.6} & \underline{38.3} & \textbf{47.5} & \textbf{39.4} \\

\bottomrule
\end{tabular}
}
\caption{Performance comparison on Physical-related multimodal benchmarks. $^{\dagger}$: Results reported in original benchmark papers. Best and second-best results of \textit{open-source} VLMs are marked in \textbf{bold} and \underline{underline}, respectively.}
\label{tab:physical-bench}
\end{table*}

\subsection{Main Results}

\noindent\textbf{SOTA open-source 7B/8B critic on physical judgment.}
On \ours-Bench (Tab.~\ref{tab:judgebench-bench}), our model attains the best overall accuracy among open-source 7B/8B models (68.0), outperforming leading general-purpose VLMs—Eagle-2.5-8B (56.0; $+12.0$), Qwen2.5-VL-7B (51.6; $+16.4$)—as well as the RL-finetuned physical models RoboBrain2.0-7B (54.7; $+13.3$) and Cosmos-R1-7B (51.1; $+16.9$).
Across sub-suites, \ours-7B delivers best or tied-best performance on AgiBot (78.8), HoloAssist (65.5; tied), and RoboVQA (86.7), while ranking second on Bridge-v2 (65.6).
Notably, its strong judgment capability also generalizes to unseen subdomains, achieving top performance on RoboFail (57.4; tied with the base Qwen2.5-VL) and LingoQA (60.0; tied best).
While physical-related RL by itself offers limited benefit to accuracy, our two-stage pipeline yields a \emph{strict} and \emph{stable} judge capable of handling embodied tasks with long-horizon, affordance-centric reasoning.

% \vspace{2mm}

\noindent\textbf{From physical critic to general-domain judging.}
As in Tab.~\ref{tab:judgebench-bench}, despite being tuned only in physical contexts, \ours generalizes well to general-domain judging scenarios.
On VL-RewardBench and Multimodal-RewardBench, it outperforms the Qwen2.5-VL base model by +4.1 and +1.9 overall, with consistent gains across general, hallucination-related, and reasoning-related judgments.
These results suggest that our physical critic training—rooted in physical perception, planning, and action evaluation—transfers effectively to broader multimodal judging across general visual domains.
% 

% \vspace{2mm}

\noindent\textbf{Improved physical reasoning as a policy.}
As shown in Tab.~\ref{tab:physical-bench}, \ours not only judges well but also strengthens \emph{policy-like} physical task handling.
On CosmosReason1-Bench, it achieves the best open-source accuracy (63.9), surpassing Cosmos-R1-7B (63.0; $+0.9$), even though the latter is trained on millions of in-domain distilled reasoning traces followed by RL finetuning.
This strong performance further generalizes to physical perception and reasoning tasks beyond the training distribution.
On CV-Bench, \ours-7B achieves the second-best average (79.7) and the best 3D score (83.9), demonstrating improved spatial grounding and cross-view consistency.
On EgoPlanBench2, \ours-7B ranks second overall (42.3) with top-two performance across all domains, highlighting robust egocentric planning capabilities.
% 

% \vspace{2mm}

\noindent\textbf{Data efficiency.}
Notably, our two-stage RL pipeline requires only 80{+}300 RL steps, with a total of 4{,}058 training samples. 
In contrast to approaches that depend on millions of supervised traces, \ours leverages \emph{verifiable} physical QA and \emph{self-referential} critic signals, enabling a more \emph{data-efficient} path toward strong critic and policy performance.

\subsection{Ablation Studies}

We further conduct two ablation studies to evaluate the impact of the two-stage RL pipeline and the proposed self-referential critic finetuning on model performance.

\begin{wraptable}{r}{0.55\textwidth}
% \begin{table}[!h]
\centering
\small
\vspace{-2mm}
\setlength{\tabcolsep}{2pt}
\scalebox{1}{
\begin{tabular}{l|cc|cc|c}
\toprule
\multirow{2}{*}{Method} 
& \multicolumn{2}{c|}{\ours-B.} 
& \multicolumn{2}{c|}{CosmosR1-B.} 
& VL-Reward. \\
\cline{2-3} \cline{4-5} \cline{6-6}
 & overall & macro & overall & macro & overall \\
% \midrule
\hline
\rowcolor{lightlightgray}
Qwen2.5-VL-7B       & 51.6 & 50.9 & 54.3  & 53.7 & 53.2  \\
- physical RL (s1)    & 53.6 & 54.6 & 61.8 & 61.2 & 52.0  \\

- physical RL (s1+s2) & 52.7 & 53.6 & 63.1 & 62.6 & 53.0  \\
- critic RL (s1+s2)      & 62.2 & 62.4 & 57.1 & 56.5 & 54.0  \\
- mixed RL (s1+s2)     & 66.7 & 68.1 & 60.2  & 59.5 & 55.5  \\
\rowcolor{lightlightblue}
- Two-stage RL (ours)         & \textbf{68.0} & \textbf{69.0} & \textbf{63.9} & \textbf{63.3} & \textbf{57.3} \\
% \midrule
\bottomrule
\end{tabular}
}
% \vspace{-1mm}
\caption{Ablation on RL strategy. “s” denotes training stage.}
\vspace{-1mm}
\label{tab:ablation-pipeline}
% \end{table}
\end{wraptable}

\noindent\textbf{Two-stage RL is necessary and complementary.} 
We compare our method against various RL strategies. In all experiments, $s1$ uses 80 RL steps and $s2$ uses 300. 
As shown in Tab.~\ref{tab:ablation-pipeline}, our two-stage pipeline achieves the best results across physical critic judgment, physical reasoning, and general-domain evaluation.
Stage~1 physical RL improves physical reasoning (+7.5 on CosmosReason1-Bench) but brings limited judgment gains (+2.0 on \ours-Bench), while Stage~2 mainly strengthens critic capability (+14.4 on \ours-Bench) and further enhances reasoning performance (+2.1 on CosmosReason1-Bench). 
Notably, the full pipeline surpasses physical RL alone in both stages on physical reasoning tasks, showing that critic training further pushes the model’s problem-solving capacities.
Overall, these results indicate that (i) Stage~1 builds foundational physical skills through policy RL warmup, and (ii) Stage~2 self-referential critic finetuning—leveraging diverse reasoning traces and high-quality preference signals—further consolidates judgment consistency, improves generalization, and reduces overfitting.

\begin{wraptable}{r}{0.51\textwidth}
% \begin{table}[!htp]
\centering
\small
\setlength{\tabcolsep}{2.2pt}
\vspace{-2mm}
\begin{tabular}{l|cc|cc|c}
\toprule
\multirow{2}{*}{Method} 
& \multicolumn{2}{c|}{\ours-B.} 
& \multicolumn{2}{c|}{CosmosR1-B.} 
& \multicolumn{1}{c}{VL-Reward.} \\
\cline{2-3}\cline{4-5}\cline{6-6}
& overall & macro & overall & macro & overall \\
\midrule
\rowcolor{lightlightblue}
\ours-7B            & \textbf{68.0}    & \textbf{69.0} & \textbf{63.9} & \textbf{63.3} & \textbf{57.3} \\
\quad - no  self-refer.   & 64.4   & 63.8  & 62.6 & 61.9 & 56.6 \\
\quad - no $r_{\mathrm{sp}}$   & 65.8    &  64.8  & 63.5 & 62.9 & 56.5 \\
\bottomrule
\end{tabular}
% \vspace{-2mm}
\caption{Ablations on self-referential critic finetuning.}
\vspace{-3mm}
\label{tab:phycrit-ablation-self-preference}
% \end{table}
\end{wraptable}

\noindent\textbf{Self-referential critic finetuning drives the gains.}
As shown in Tab.~\ref{tab:phycrit-ablation-self-preference}, removing the explicit self-referential process during finetuning (\emph{no self-refer.}) reduces \ours-Bench performance from 68.0 to 64.4 ($-3.6$).
Keeping the self-reference prompt but removing the self-prediction reward $R_{\mathrm{sp}}$ leads to a smaller yet noticeable drop to 65.8 ($-2.2$). 
Similar trends appear on CosmosR1-Bench (63.9 $\rightarrow$ 62.6 / 63.5) and VL-Reward (57.3 $\rightarrow$ 56.6 / 56.5). 
By explicitly instructing the judge model to ground its evaluations in its own reasoning and problem-solving behavior, and by applying rewards that reinforce accurate self-reasoning, the model simultaneously learns to produce \emph{generalized, consistent judgment} and \emph{faithful reasoning}.

\subsection{Analysis}

\noindent\textbf{Better self-predictions $\Rightarrow$ better judgments.}
To assess how self-prediction accuracy affects judgment quality, we conduct chi-square tests between the correctness of \ours’s own answer and its downstream judgment in the self-referential critic process. The Stage~1 model with physical RL warmup already shows a strong positive association ($\chi^2=51.07$, $p=8.93\times10^{-13}$), and self-referential critic finetuning further strengthens this dependence in the final model ($\chi^2=161.76$, $p=4.66\times10^{-37}$). These results support the premise that a strong physics-aware critic should “solve before judging,” using its own grounded prediction to avoid spurious correlations and unsupported verdicts.
% 

% \vspace{2mm}

\noindent\textbf{Limitations.} While self-referential critic fine-tuning has proven more effective than conventional pairwise critic training, it additionally requires ground-truth answers for the multimodal prompts. Although actions and planning can often be verified through physical-world feedback, this requirement limits its applicability to fully open-ended scenarios. Future work may explore self-verification or meta-judging strategies to replace the explicit accuracy reward used for self-prediction. Furthermore, \ours could be extended toward multi-round critic self-refinement, leveraging critic signals to iteratively improve self-generation.
\section{Conclusion}
We presented \ours, a multimodal critic model tailored for physical AI scenarios involving perception, causal reasoning, and action or plan evaluation. By introducing a two-stage RLVR pipeline with \emph{self-referential critic finetuning}, \ours grounds its judgments in its own physically informed predictions, yielding more consistent, interpretable evaluations and stronger physical reasoning. We further introduced \ours-Bench, a challenging benchmark for physical-domain multimodal judging. Experiments show that \ours achieves the best performance among open-source 7B/8B models on physical judgment, generalizes well to broader multimodal reward tasks, and enhances physical reasoning. Together with our analysis, \ours paves the way toward more reliable, physical-aware multimodal AI evaluation systems.

\clearpage
\appendix

% \clearpage
% \appendix

\section{Additional Results and Analysis}

In this section, we first analyze the robustness of (i) reward weight hyperparameters and (ii) critic prompt design; we then validate \ours as a reward for (iii) test-time scaling and (iv) downstream policy training, and finally compare it with (v) a general-domain multimodal critic.

\begin{wraptable}{r}{0.55\textwidth}
% \begin{table}[!hp]
\centering
% \small
% \setlength{\tabcolsep}{2.5pt}
\renewcommand{\arraystretch}{0.95}
\vspace{-3mm}
\begin{tabular}{l|cc|cc|c}
\toprule
\multirow{2}{*}{$\alpha_{sp}$} 
& \multicolumn{2}{c|}{\ours-B.} 
& \multicolumn{2}{c|}{CosmosR1-B.} 
& \multicolumn{1}{c}{VL-Reward.} \\
\cline{2-3}\cline{4-5}\cline{6-6}
& overall & macro & overall & macro & overall \\
\midrule
0.1           & 65.8 & 66.1 & 62.9 & 62.3 & \textbf{58.6}  \\
% \rowcolor{lightlightblue}
0.2         & \textbf{68.0}    & \textbf{69.0} & \textbf{63.9} & \textbf{63.3} & 57.3 \\
0.3           & 66.2 & 66.4 & 63.3 & 62.7 & 57.2 \\
0.4           & 65.3 & 65.2 & 63.7 & 63.1 & 56.4 \\
\bottomrule
\end{tabular}
\renewcommand{\arraystretch}{1.0}
\vspace{-1mm}
% \begin{center}
\caption{Ablations on the self-prediction reward weight ($\alpha_{sp}$).}
% % \end{center}
\vspace{-2mm}
\label{tab:phycrit-ablation-self-pred-alpha-appendix}
% \end{table}
\end{wraptable}

\vspace{-0.5mm}
\textbf{Ablations on $\alpha_{sp}$.} We fix the format reward weight at $\alpha_{form}=0.1$ and vary the self-prediction reward weight $\alpha_{sp}$ while setting the critic reward weight to $\alpha_{crit}=0.9-\alpha_{sp}$. As shown in Tab.~\ref{tab:phycrit-ablation-self-pred-alpha-appendix}, \ours exhibits relatively robust performance across different choices of $\alpha_{sp}$. We adopt $\alpha_{sp}=0.2$ in our final configuration, as it provides the best overall balance across judge and reasoning benchmarks.

\begin{wraptable}{r}{0.54\textwidth}
% \begin{table}[!htp]
\centering
% \small
\setlength{\tabcolsep}{2pt}
\vspace{-3mm}
\renewcommand{\arraystretch}{0.95}
\begin{tabular}{l|cc|cc|c}
\toprule
\multirow{2}{*}{Critic Prompt} 
& \multicolumn{2}{c|}{\ours-B.} 
& \multicolumn{2}{c|}{CosmosR1-B.} 
& \multicolumn{1}{c}{VL-Reward.} \\
\cline{2-3}\cline{4-5}\cline{6-6}
& overall & macro & overall & macro & overall \\
\midrule
\rowcolor{lightlightblue}
Full prompt            & \textbf{68.0}    & \textbf{69.0} & \textbf{63.9} & \textbf{63.3} & \textbf{57.3} \\
\quad - no  criteria   & 63.9   & 64.8  & 62.0 & 61.3 & 55.1 \\
\bottomrule
\end{tabular}
\renewcommand{\arraystretch}{1}
\vspace{-1.5mm}
\caption{Ablations on critic prompt criteria.}
\vspace{-1mm}
\label{tab:phycrit-ablation-critic-prompt-criteria-appendix}
% \end{table}
\end{wraptable}

\vspace{-0.5mm}
\textbf{Ablations on critic prompts.}
As shown in Tab.~\ref{tab:critic_prompt_self_reference}, our critic prompts include detailed evaluation criteria that guide the model’s rollout during the critic process—for example, truthfulness, visual groundedness, logical validity, and efficiency for reasoning assessments.
To assess the importance of these hand-crafted criteria, we conduct an ablation study that removes all criteria and report the results in Tab.~\ref{tab:phycrit-ablation-critic-prompt-criteria-appendix}. Removing the criteria leads to clear drops in judgment accuracy, physical reasoning performance, and generalization to image-domain judgments.
Since our GRPO objective relies only on the self-prediction reward $r_{sp}$ and critic accuracy reward $r_{crit}$, the criteria in the critic prompts become crucial for shaping and regularizing \ours’s critic reasoning behavior during reinforcement finetuning.

\vspace{-1mm}
\noindent This finding contrasts with prior work such as \cite{wang2025llava}, which uses a simple pairwise critic prompt for general-domain critic training. Our results indicate that building reliable critic capabilities for physical-AI tasks is considerably more challenging and requires explicit, structured critic guidance.

\begin{wrapfigure}{r}{0.5\textwidth}
    \centering
    \vspace{-2.5mm}
    \includegraphics[width=1.0\linewidth]{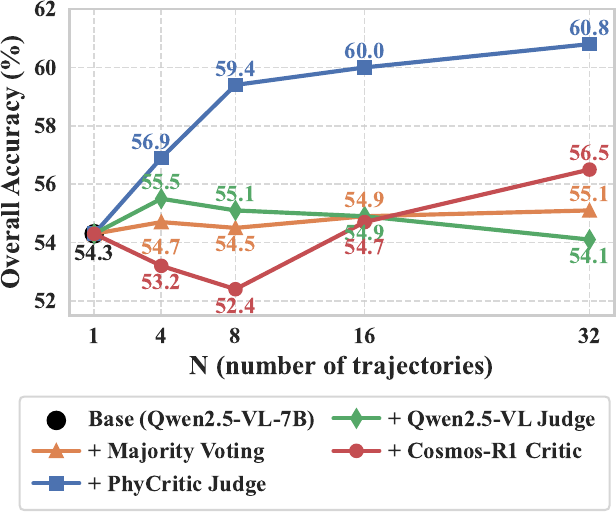}
    % \vspace{-5.5mm}
    \vspace{-3.5mm}
    \caption{Comparison of Best-of-$N$ ensemble mechanisms on CosmosReason1-Bench. Using \ours-7B as the judge consistently improves the base Qwen2.5-VL-7B-Instruct model.}
    \vspace{-10mm}
    \label{fig:phycrit-bon-appendix}
\end{wrapfigure}

% \vspace{-0.5mm}
\textbf{\ours for best-of-N sampling.} Here, we evaluate the effectiveness of \ours as a test-time scaling judge for best-of-$N$ sampling. We first employ Qwen2.5-VL-7B-Instruct as the \textit{policy model} to generate $N$ candidate reasoning trajectories for each question in CosmosReason1-Bench~\cite{azzolini2025cosmos} under a thinking prompt with a temperature of $\tau=0.6$. \ours-7B is then applied through a \textit{pairwise knockout} procedure: starting from the first two trajectories, \ours judges the pair and advances the preferred response to compete with the next candidate. This process continues for $N-1$ rounds until a final winner is selected. We compare this strategy against three baselines:
1) majority voting over the final predicted answers;
2) using Qwen2.5-VL-7B-Instruct~\cite{bai2025qwen2} itself as the judge for pairwise knockout; and 3) using the physical RL-finetuned Cosmos-Reason1-7B~\cite{azzolini2025cosmos} as the judge.

As shown in Fig.~\ref{fig:phycrit-bon-appendix}, applying \ours to select the best reasoning trajectory yields the most consistent performance gains as $N$ increases, achieving a $+6.5$-point improvement  at $N=32$ for the base model on CosmosReason1-Bench (60.8 vs 54.3). 
In contrast, using the physically finetuned Cosmos-Reason1-7B or the base Qwen2.5-VL models as the critic yield much smaller or even no improvements, reflecting their limited physical-domain judgment capacity. By reliably identifying the high-quality trajectory among multiple candidates, \ours serves as an effective ensemble mechanism for enhancing test-time performance.

\textbf{\ours for downstream policy training.}
To validate \ours as a reward signal for guiding downstream policy training, we apply \ours within a self-improving DPO framework. Starting from Qwen2.5-VL-7B-Instruct, we generate 8 responses per prompt on Cosmos-RL data, use \ours-7B to score all ordered response pairs and select best–worst responses as preference pairs for DPO training (learning rate $1\mathrm{e}{-}6$, 5 epochs). The resulting DPO-trained policy is evaluated on CosmosReason1-Bench.
As shown in Tab.~\ref{tab:phycritic-dpo}, \ours-guided DPO achieves substantial gains over the base Qwen2.5-VL model and consistently outperforms an answer-verifier baseline (random correct–incorrect pairs), indicating that \ours provides an effective reward signal beyond answer correctness for improving downstream physical reasoning. %

\begin{table}[!h]
    \centering
    % \small
    \vspace{-1mm}
    % \renewcommand{\arraystretch}{0.9}
    % % \scalebox{0.85}{
    % \setlength{\tabcolsep}{3pt}
    \begin{tabular}{l|ccccccc}
    \toprule
        Model &  overall & AgiBot & HoloAssist & RoboVQA & BridgeV2 & RoboFail \\
    \midrule
       \rowcolor{lightlightgray} Qwen2.5-VL-7B-Instruct  & 54.3 & 40.4 & 50.0 & 82.5 & 38.0 & 57.6  \\
       + answer-verify DPO & 57.5 & 39.0 &  60.0 & 86.4 & 39.0 & 60.0 \\
       + \ours-7B DPO & 60.0&  41.0 & 66.0 & 90.0 &  39.0 & 61.0 \\
    \bottomrule
    \end{tabular}
    % }
    % \vspace{-3.7mm}
    \renewcommand{\arraystretch}{1.0}
    \vspace{-1mm}
    \caption{\ours for guiding DPO training on Qwen2.5-VL-7B-Instruct, evaluated on CosmosReason1-Bench.}
    % % \vspace{-5mm}
    \label{tab:phycritic-dpo}
\end{table}

\textbf{Comparison with a general-domain critic.} We compare \ours with UnifiedReward-Think~\cite{wang2025unifiedthink}, a recent multimodal reward model fine-tuned with RL on general understanding and generation preference data.
As shown in Tab.~\ref{tab:unifiedreward-think}, UnifiedReward-Think performs strongly on general-domain reward benchmarks, but shows only marginal gains on physical-domain judgment and reduced physical reasoning performance. In contrast, \ours achieves robust gains on physical-domain judgment and reasoning, and its critic capacity transfers to general-domain judging, despite not being trained on general-domain critic data.

% \textcolor{setting}{clear}.

\begin{table}[!h]
    \centering
    % \small
    % \vspace{-3.2mm}
    % \renewcommand{\arraystretch}{0.88}
    % % \scalebox{0.85}{
    \setlength{\tabcolsep}{3pt}
    \vspace{-1mm}
    \begin{tabular}{l|cc|cc}
    \toprule
        Model & \ours-B. & CosmosReason1-B. & VL-RewardB. & Multimodal-RewardB.  \\
    \midrule
       \rowcolor{lightlightgray} Qwen2.5-VL-7B-Instruct & 51.6 & 54.3 & 53.2 & 64.0  \\
       UnifiedReward-Think-Qwen-7B &  52.4 & 51.8 & 73.8 & 66.1 \\
       \ours-7B & 68.0 & 63.9 & 57.3 & 65.9 \\
    \bottomrule
    \end{tabular}
    % }
    % \renewcommand{\arraystretch}{1.0}
    \vspace{-1mm}
    \caption{Comparison with UnifiedReward-Think. Both models are initialized from Qwen2.5-VL-7B-Instruct.}
    % \vspace{-4.2mm}
    \label{tab:unifiedreward-think}
\end{table}

% \clearpage
\clearpage

\section{Qualitative Examples}

In this section, we provide additional case studies to further compare the judging and physical reasoning behaviors of \ours with those of baseline models.

\begin{itemize}[itemsep=4pt]
    \item Table~\ref{tab:example_phycrit_phycrit-bench} provides an example of \ours judging a pair of model responses for physical-related evaluation tasks via the self-referential critic process. By first generating its own reasoning to solve the problem and then explicitly referencing this reasoning during critique, \ours can accurately distinguish differences in the quality of physical-related reasoning rather than relying on shallow stylistic cues, leading to more consistent judgments supported by grounded and clear justifications.

    \item Table~\ref{tab:example_phycrit_cosmos} illustrates a case where \ours is used as a \textit{policy model} for physical reasoning. The self-referential critic training process further strengthens its physical perception and reasoning capabilities, enabling \ours to generate visually grounded and internally consistent reasoning that leads to reliable final answers.

    \item Table~\ref{tab:example_phycrit_vlreward} presents an example of \ours judging a pair of model responses in a general image domain. Although \ours is trained exclusively on physical-related critic data, its physically grounded critic capability transfers effectively to broader visual tasks, enabling reliable and visually grounded judgments even when instructed to use different critic formats and to evaluate general image-based model responses.
\end{itemize}

\begin{table*}[!htbp]
\begin{minipage}{1.0\textwidth}
\centering
\small
\scalebox{0.91}{
    \begin{tabular}{p{2cm} p{15cm}}
    \toprule
     \multicolumn{2}{l}{\bf \ours in Judging Physical-Related Domains}  \\
    \midrule
     & \includegraphics[height=3cm]{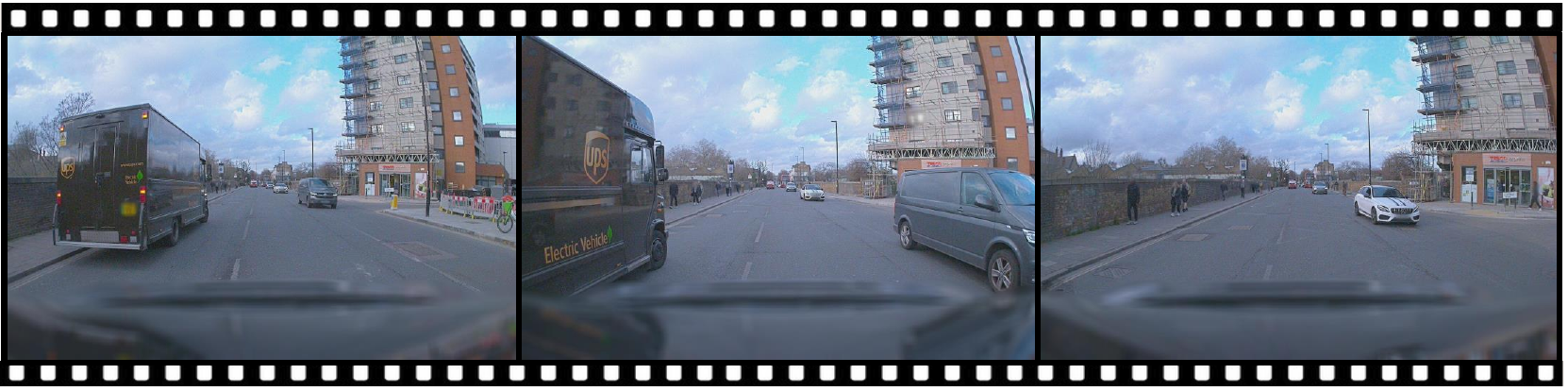} \\
    Question & What should you do next in this driving scenario? A. Continue driving as the truck is stopped and you have the right of way. B. Stop and wait for the truck to move first. C. Reverse to allow the truck to proceed. D. Overtake the truck by entering the opposite lane. \\
    Response 1 & \texttt{<think>} 
    In this scenario, the vehicles are on a two-lane road with traffic in both directions. The UPS truck appears to be stationary or moving very slowly, creating a potential obstruction. Although the reason for the stop is unclear, it is blocking the left lane.
    Option A is unsafe since the truck may be stopped legitimately, and entering its lane could cause a collision. Option B safely checks that the road is clear before moving. Option C is dangerous due to possible traffic behind you. Option D is risky because of oncoming traffic. 
    \texttt{</think>}  \texttt{<answer>  B </answer>} \\
    Response 2 & \texttt{<think>}  In this driving scenario, the UPS truck appears to be preparing to turn or has stopped temporarily, possibly to yield to oncoming traffic. However, the road ahead is clear, and there seems to be no immediate obstacle preventing the driver from continuing to drive forward safely. Since the truck is currently stationary and there is no sign indicating that the vehicle must stop, it suggests that the driver has the right of way. \texttt {</think>}  \texttt{<answer> A </answer>} \\
    \midrule
   %  \makecell[l]{Qwen2.5-VL-7B \\ -Instruct}
   Ground Truth & Response 2  $>$ Response 1 \\
   \midrule
   Qwen2.5-VL-7B-Instruct
     &  \#\#\#  Response 1:  - **Reasoning Process Quality**:    
     - Truthfulness: The response correctly identifies that the UPS truck is stationary or moving slowly and may obstruct the lane.  
     - Visual Groundedness: It references the image showing the UPS truck on the left side, partially blocking the left lane. 
     - Logical Validity: The reasoning is sound—assessing the truck’s movement, the obstruction risk, and the safest action. 
     - Efficiency and Clarity: The response is clear and concise, avoids repetition, and presents a coherent flow of reasoning.  
     - **Final Answer Accuracy**: \textcolor{brown}{The final answer is accurate; stopping and waiting for the truck to move first is indeed a safe and responsible action.} \\
    & \#\#\# Response 2:  - **Reasoning Process Quality**:    
    - Truthfulness: The response correctly observes that the truck appears stationary or preparing to turn and notes no immediate obstacle.   
    - Visual Groundedness: It references the visual evidence where the truck is visible on the left side 
    and no traffic ahead obstructs the lane.  
    - Logical Validity: The reasoning is logical, given the truck’s stationary state and the unobstructed lane, and the conclusion about having the right of way is reasonable.
     - Efficiency and Clarity: The response is clear but \textcolor{brown}{slightly less structured than Response 1, as it does not explicitly outline all safety-checking steps.} 
    - **Final Answer Accuracy**: The final answer is accurate; continuing to drive as the truck is stopped and there is no immediate danger is safe and efficient.  \\
    & \#\#\# Evaluation Summary:  Both responses demonstrate factual accuracy, strong visual grounding, a coherent logical flow, and clear, efficient communication. However, **Response 1** is \textcolor{brown}{more comprehensive and structured, clearly detailing the steps for assessing the situation and identifying the safest action, which makes its reasoning easier to follow.}   \texttt{Decision: Response 1 is better} \\
    \midrule

    \ours-7B & \texttt{<pred\_think>}  The image shows a busy street with a UPS truck in the left lane, seemingly stopped or moving slowly. There are other cars and pedestrians on the sidewalks. \textcolor{blue}{Given the situation, the truck is not obstructing the right lane where the driver's car is positioned, allowing for a safe passage.} 
    The road ahead is clear, and there is no visible sign that necessitates stopping. Therefore, the most logical action is to continue driving, assuming the truck will eventually move out of the way.  \texttt{</pred\_think>} \texttt{<pred>}  A \texttt{</pred>} 
     \texttt{<think>} Both Model Responses 1 and 2 acknowledge that the truck is stationary and there is no immediate danger from the truck itself, allowing the driver to continue driving. However, they differ in their assessment of the safety and appropriateness of waiting versus continuing.  Model Response 1 emphasizes the safety aspect by advising to stop and wait for the truck to move first, citing the possibility of the truck being stopped for a legitimate reason. This response is \textcolor{red}{grounded in the visual evidence that the truck is stationary} and not obstructing the path, but \textcolor{red}{it slightly overemphasizes the need to wait due to the lack of visible signs or indications that stopping is necessary.}    Model Response 2 \textcolor{red}{correctly identifies that the truck is stationary and does not present an obstacle, allowing the driver to continue driving} without waiting. This approach aligns well with the visual evidence and the logic of the situation. The road ahead is clear, and the truck is not obstructing the right lane, making it safe to continue driving.  My reasoning process aligns more closely with Model Response 2. 
     \texttt{</think>}  \texttt{\textbackslash boxed\{Response 2 is better\}} \\

    \bottomrule
    \end{tabular}
}
\captionof{table}{Example on \ours-Bench. Qwen2.5-VL-7B focuses on \textcolor{brown}{superficial structures} and produces \textcolor{brown}{uninformative critiques}. In contrast, \ours delivers more aligned judgments by first \textcolor{blue}{identifying the truck’s status during its own reasoning} and then grounding this reasoning in the critic process, allowing it to accurately \textcolor{red}{capture the quality differences} between the responses.}
\label{tab:example_phycrit_phycrit-bench}  
\end{minipage}
\end{table*}

\begin{table*}[!htbp]
\begin{minipage}{1.0\textwidth}
\centering
\scalebox{0.96}{
    \begin{tabular}{l p{13cm}}
    \toprule
     \multicolumn{2}{l}{\bf \ours in Physical Reasoning}  \\
    \midrule
     & \includegraphics[height=3.2cm]{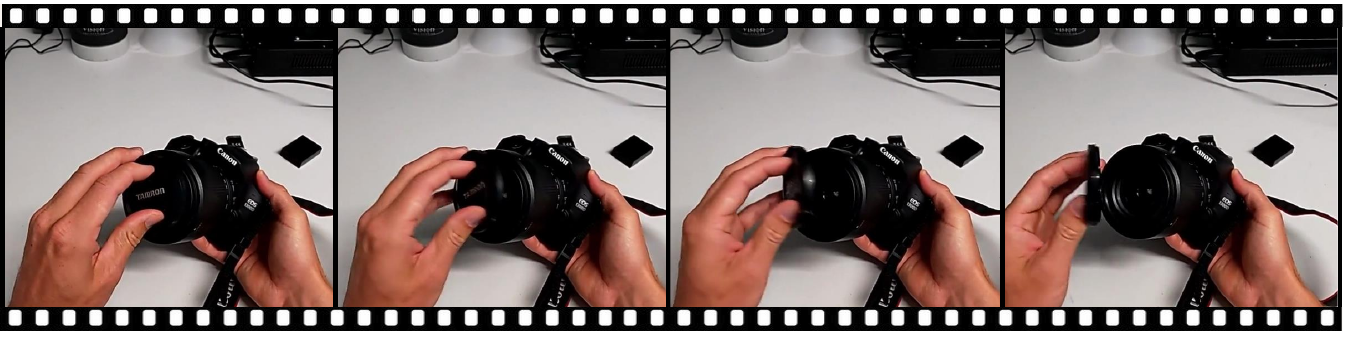} \\
    Question & The overall goal is ``The student detaches the cover.". The agent in the video is currently performing one subtask out of many to complete this goal. For the agent in the video, what is the most plausible next immediate subtask from the choices below?\\
    & A: grab lens\_cover \\
    & B: grab dslr \\
    & C: lift dslr \\
    & D: place lens\_cover\\
    \midrule
    Answer & D\\
    \midrule
    Cosmos-Reason1-7B & \texttt{<think>} Okay, let's see. The user provided  the video and the current action. The main goal is for the student to detach the cover. \textcolor{brown}{The agent is currently holding the camera with both hands, preparing to remove the lens cap. So, the current step is that the left hand is adjusting the lens cover. The next logical step after adjusting would be to actually take it off.} Since the lens cap is already being handled, the next action should involve removing it. The options are A (grab lens\_cover), B (grab dslr), C (lift dslr), D (place lens\_cover). Option A makes sense because once you've adjusted or positioned the lens cap correctly, the next step is to grab it to remove it. Option B and C involve grabbing or lifting the DSLR, which might not be necessary if the goal is just to detach the cover. Option D suggests placing the lens cover somewhere else, but since the task is detachment, placing isn't the immediate next step. So the most plausible is A.\texttt{</think>} \\
    & \texttt{<answer>} A \texttt{</answer>} \\
    \midrule
    \ours-7B & \texttt{<think>} The agent is holding a camera with a lens cover attached. The goal is to detach the cover. \textcolor{red}{The agent has their hand on the lens cover, which suggests they are removing it. The most logical next step would be to place the lens cover down after detaching it.} - Option A (grab lens\_cover) seems redundant since the lens cover is already being held. - Option B (grab dslr) does not make sense because the agent should not be grabbing the camera again at this point. - Option C (lift dslr) is incorrect because lifting the DSLR is not necessary to detach the lens cover. - Option D (place lens\_cover) aligns with the action needed to properly detach the lens cover. Therefore, the most plausible next immediate subtask is to place the lens cover. \texttt{</think>} \\
    & \texttt{\textbackslash boxed\{D\}} \\
    \bottomrule
    \end{tabular}
}
\captionof{table}{Example of \ours on CosmosReason1-Bench~\cite{azzolini2025cosmos}. 
For \textit{policy-like} problem solving on physical reasoning tasks, \ours performs more visual-grounded, logically consistent, and efficient reasoning to infer the most plausible next subtask, without introducing \textcolor{brown}{stepwise contradictions.}}
\label{tab:example_phycrit_cosmos}  
\end{minipage}
\end{table*}

\begin{table*}[!htbp]
\begin{minipage}{1.0\textwidth}
\centering
\scalebox{1}{
    \begin{tabular}{p{3cm} p{12.2cm}}
    \toprule
     \multicolumn{2}{l}{\bf \ours in Judging General Visual Domains (Out of Training Distribution) }  \\
    \midrule
     & \includegraphics[height=4cm]{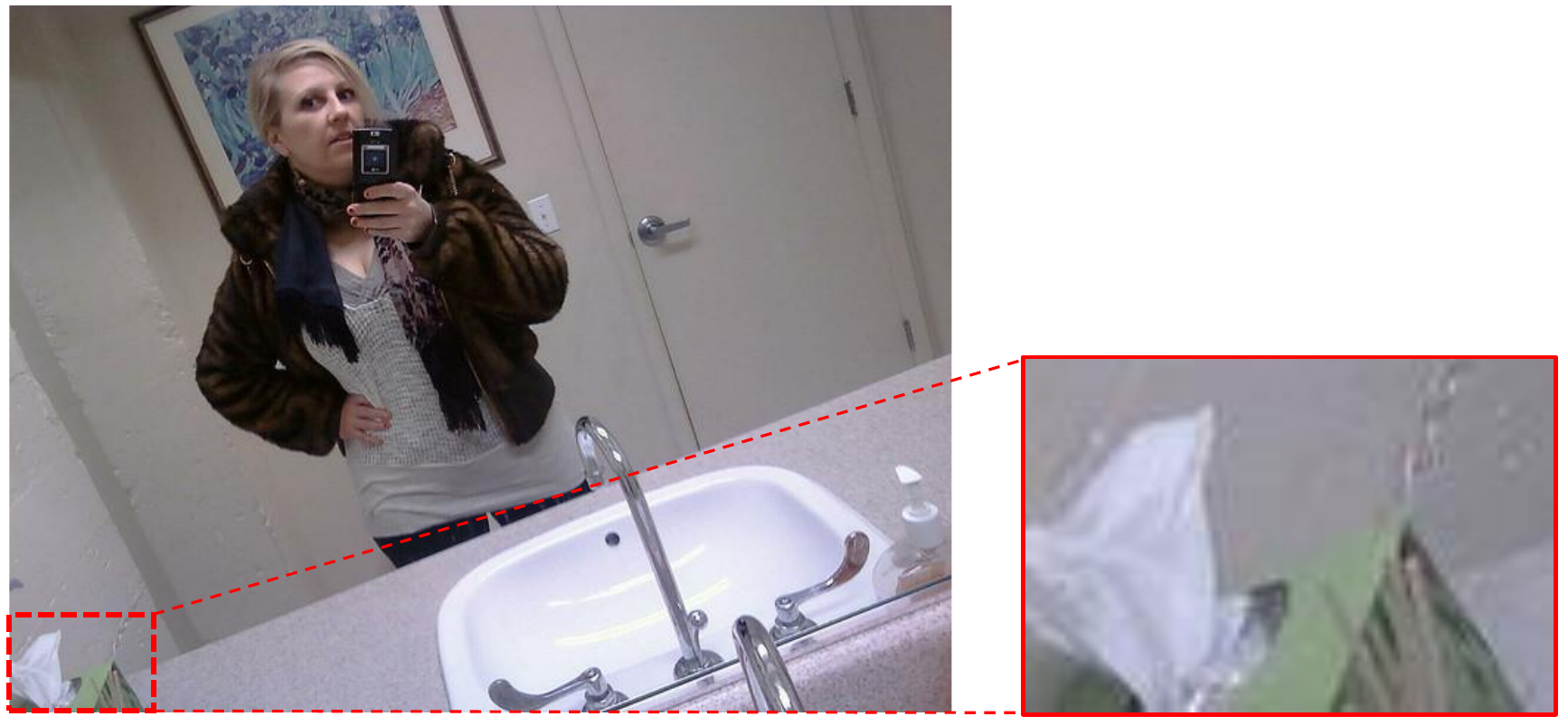} \\
    Question & What color is the tissue box on the counter?\\
    
    Response 1 & The tissue box on the counter in the picture is green.\\
    Response 2 & The tissue box on the counter in the picture is white.\\
    \midrule
    Ground Truth & Response 1 $>$ Response 2 \\
    \midrule
    % \makecell[l]{Qwen2.5-VL-7B \\ -Instruct} 
    Qwen2.5-VL-7B-Instruct
    & 1. **Accuracy**:   - Both responses directly address the color of the tissue box mentioned in the image query. However, only one of them can be correct based on what is visible in the image.  \textcolor{brown}{ - Based on the visual evidence in the image, there is no green present, confirming that Response 1 is not accurate.   - The tissue box is indeed white in the image.} \\
    & 2. **Completeness**:   - Both responses provide full responses to the question by naming a single color. Complete responses are typically expected to include only relevant responses without additional unnecessary text. Both Response 1 and Response 2 do this well as they solely focus on the color. \\
    & 3. **Clarity**:   - Both responses are straightforward and unambiguous, making it clear what color the tissue box is.  - There is no extraneous information, ensuring clarity for the reader. \\
    & 4. **Relevance**:  - Both responses directly relate to the question and the image by providing the color of a specific object (the tissue box) mentioned in the query.   - Neither response includes unrelated details, thus being relevant. \\
    & **Overall Judgment**: After evaluating the accuracy, completeness, clarity, and relevance, Response 2 is better. \\
    \midrule
    \ours-7B & **1. Accuracy:** - **Response 1:** The tissue box on the counter appears green.\ **Response 2:** The tissue box on the counter appears white. \\
    & **2. Completeness:** - Both responses provide a straightforward response to the question about the color of the tissue box. However, they do not offer additional information that might not be relevant to the question, such as the location or condition of the tissue box. \\
    & **3. Clarity:** - Both responses are clear and easy to understand. They directly response the question without ambiguity. \\
    & **4. Relevance:** - Both responses directly relate to the question by addressing the color of the tissue box in the picture. \\
    & **Overall Judgment:** - \textcolor{red}{Response 1 is better because the color of the tissue box in the image is green, whereas Response 2 states it is white, which is incorrect.}  \#\#\# Detailed Response: - \textcolor{blue}{The tissue box on the counter is noticeably green, as seen in the image. It appears compact and situated near the sink. The surrounding surfaces are neutral in color, and the green hue of the tissue box contrasts with the lighter colors of the countertop and the sink.}\\
    & Final Response: Response 1 is better. \\
    \bottomrule
    \end{tabular}
}
\captionof{table}{Example of \ours on VL-RewardBench~\cite{lambert2024rewardbench}. Due to improved \textcolor{blue}{physical perception}, \ours accurately identifies the tissue box as green and avoids \textcolor{brown}{hallucinated colors}, producing \textcolor{red}{judgments that better align with human preference.} Note that neither image-domain critic data nor the critic prompts used in VL-RewardBench were seen during \ours' training. }
\label{tab:example_phycrit_vlreward}  
\end{minipage}
\end{table*}

\clearpage
\setcitestyle{numbers}
\bibliographystyle{plainnat}
\bibliography{main}

\end{document}